# A robust and compliant robotic assembly control strategy for batch precision assembly task with uncertain fit types and fit amounts

**Bin Wang, Jiwen Zhang, Song Wang, Dan Wu***

*State Key Laboratory of Tribology in Advanced Equipment, Beijing Key Laboratory of Precision/Ultra-Precision Manufacturing Equipment Control, Department of Mechanical Engineering, Tsinghua University, Beijing, China*

ABSTRACT

In some high-precision industrial applications, robots are deployed to perform precision assembly tasks on mass batches of manufactured pegs and holes. If the peg and hole are designed with transition fit, machining errors may lead to either a clearance or an interference fit for a specific pair of components, with uncertain fit amounts. This paper focuses on the robotic batch precision assembly task involving components with uncertain fit types and fit amounts, and proposes an efficient methodology to construct the robust and compliant assembly control strategy. Specifically, the batch precision assembly task is decomposed into multiple deterministic subtasks, and a force-vision fusion controller-driven reinforcement learning method and a multi-task reinforcement learning training method (FVFC-MTRL) are proposed to jointly learn multiple compliance control strategies for these subtasks. Subsequently, the multi-teacher policy distillation approach is designed to integrate multiple trained strategies into a unified student network, thereby establishing a robust control strategy. Real-world experiments demonstrate that the proposed method successfully constructs the robust control strategy for high-precision assembly task with different fit types and fit amounts. Moreover, the MTRL framework significantly improves training efficiency, and the final developed control strategy achieves superior force compliance and higher success rate compared with many existing methods.

## 1. Introduction

With the development of intelligent manufacturing, deploying robots to replace manual operations in peg-in-hole assembly tasks has greatly enhanced production efficiency and product quality [1]. In the batch assembly process of 3C (computers, communication and consumer electronics) products, high-precision assembly tasks are more challenging as the mating components to be assembled are in tight clearance fits (0.01 mm) or even small interference fits [2]. In high-precision industrial scenarios, the fit types of numerous mating components are designed as transition fits, such as mobile phone lenses. To enhance assembly efficiency, the geometric dimensions of the mating components are typically not individually measured prior to the assembly. As a result, the unavoidable manufacturing errors of the mating components lead to uncertain fit types and fit amounts even in the same batch of manufactured pegs and holes. Distinct fitting states correspond to significantly different mechanical properties during the assembly process. This requires the robot assembly controller must possess sufficient robustness to effectively handle uncertainties of the fit states. Otherwise, robotic solutions cannot be reliably deployed in industrial production environments. This paper focuses on constructing a robust and compliant robotic assembly control strategy to accommodate uncertainties of fit types and fit amounts in the batch precision assembly tasks, which is of great significance for urgent assembly automation requirements of 3C products.

Currently, there are relatively few researches focusing on the robustness in robotic batch precision assembly, but precision assembly control methods have been extensively studied. According to the different design principles, precision assembly control methods can be divided into two categories: model-based control methods and learning-based control methods [3].

Model-based control methods aim to describe the relationship between the peg-in-hole pose error and the sensor signals (such as force, vision and tactile signals) through mathematical modeling, and then use it to optimize the design of feedback controllers. By using the vision information, visual servo control methods [4, 5] can intuitively analyze the pose error between the peg and hole for high-precision alignment, but it is always time-consuming and lacks effective control of the contact force during the insertion process. Force-based control methods achieve higher compliance. Based on jamming analysis [6], Gaussian model [7], mass-spring-damper model [8] and elastic mechanics [9], researchers construct the mathematical models of the contact force and moment between the peg and hole. Then the derived model is used to guide the modification or dynamic adjustment of the PD controller [6, 7] or admittance controller [8-10]. However, model-based force controller cannot achieve satisfied performance because the modeling is over simplified, especially for complex-shaped objects. In some special high-precision assembly scenarios (for example, the robot grasps the peg through a flexible spring [11] or suction cup [12]), force and vision information can be fused simultaneously into the controller to achieve both smaller contact force and higher assembly speed. All the above model-based control methods are manually designed and usually not robust for batch precision assembly tasks. The change of fit types and amounts of the mating parts will lead to the change of the contact force model, so the controller parameters need to be manually re-designed.

* *Corresponding author*
E-mail address: wud@mail.tsinghua.edu.cn (D. Wu).

Learning-based control methods depend on the data rather than theoretical models to construct the controller, so they can be applied to components with complex shapes. Methods based on the demonstration learning [13] and the supervised learning [14] are essentially fitting expert's teaching trajectories or pre-labeled datasets. For precision assembly tasks, data acquisition is difficult and data quality cannot be guaranteed, which constrains the performance of the learned controller. In contrast, reinforcement learning (RL) can continuously acquire data and improve policies through online interaction with the environment to achieve optimal control. To improve the efficiency and stability in training the assembly strategies, RL are often used to dynamically adjust the pre-defined force controller parameters [15-18] or output the residual actions [19, 20], instead of letting the agent to learn from random actions [21]. These methods can eventually obtain the probabilistically optimal control strategy, even for components with complex shapes. However, the above RL-based control method is only effective for the specific mating component used for training, and exhibit limited robustness under varying conditions. When the shape, size, or fit type of the components change, the control performance significantly decreases, so retraining or transfer learning are required [22].

In assembly tasks with low precision (0.5mm~2.0mm clearance fit), some researches have explored constructing robust RL strategies for components with various shapes. Most studies use both force and visual signals as inputs, and the visual signals essentially provide the adaptability to large initial alignment errors. The most direct idea is to perform domain randomization in simulation assembly environment and then transfer learned strategy for real-world application [23, 24]. The sim-to-real gap for visual perception is relatively small, whereas a significant sim-to-real discrepancy exists in contact force dynamics. Therefore, sim-to-real method is efficient only for assembly tasks with large clearance fit ( > 1.0 mm). In real-world training scenarios, Yun [26] unifies the visual and dynamic state representation of assembly tasks for different electrical connectors through pre-trained encoders and then construct a uniform strategy, but the state encoder is separately trained for all connecters and cannot be transferred to new untrained connectors. Utilizing only force signals, Jin [25] concentrates on the insertion process and develops a hybrid ensemble learning approach for multiple peg-in-hole assembly tasks with different shapes, but numerous agents are trained separated, which is very inefficient. In summary, these researches usually only pursue improving the success rate without evaluating the contact force compliance during assembly, which is not sufficient for high-precision industrial tasks.

To the best of our knowledge, few research has focused on the robust RL control strategies for uncertain fit types and fit amounts in batch precision assembly. The challenge arises from the fact that the actual fit amount is a continuously varied and unknown parameter within a certain range during high precision batch assembly, while the shapes in the previous studies [23-26] are several known categories that can be traversed for trained. In addition to the success rate, the force compliance during the control process should also be considered for avoiding damage to parts, especially for assembly with interference fit. Constructing a robust and compliant control strategy has emerged as a critical challenge.

In the field of general artificial intelligence, multi-task reinforcement learning (MTRL) [27] can capture the common information between tasks to improve the policy's robustness by training multiple sets of policies simultaneously. Policy distillation [28] integrates multiple teacher models into one student model to obtain stronger and more robust capabilities. These techniques are potential to be combined with the compliance control methods to obtain robustness in batch precision assembly tasks.

In this study, we propose an efficient construction method of the robust and compliant control strategy for robotic batch precision assembly tasks. The overall idea includes task decomposition, subtasks strategies learning and strategies integration. First, the batch precision assembly task is decomposed into several subtasks, and the fit type and fit amount in each subtask is different but deterministic. Subsequently, considering inherent similarities between subtasks, control strategies for different subtasks are simultaneously and efficiently learned based on a MTRL training framework. During this process, a force-vision fusion controller is combined with RL to further improve the assembly compliance. Finally, the multi-teacher policy distillation is utilized to obtain a robust control strategy by sufficiently leveraging the subtasks strategies and training experience. Although the discrete fit amounts in several subtasks cannot cover the continuously varying range, MTRL can implicitly learn robust representations across subtasks, and policy distillation can integrate the capabilities of all strategies, contributing to the great performance when dealing with unknown fit amounts. The effectiveness of the proposed method is verified in real-world experiments. For batch precision assembly of irregular hexagonal components with varying fit amounts between 0.02mm interference fit to 0.02mm clearance fit, a robust and compliant control strategy is successfully constructed, achieving high success rate and small contact force.

It should be noted that this paper focuses on batch precision assembly tasks of components with given shape but uncertain fitting types and fitting amounts. From a broader perspective, the proposed method can also be adopted to construct a robust assembly control strategy for components with multiple shapes, simply by modifying the subtasks decomposition pattern. This is validated through simulation studies presented in the Supplementary Materials, further demonstrating the effectiveness and practical value of the proposed approach.

The rest of the paper is organized as follows. Section 2 gives a brief introduction about the robotic batch precision assembly scenario and the method framework. Section 3 develops the construct method of the robust and compliant control strategy. Section 4 provides the real-world experiment verification results. Section 5 gives the summarization of this paper.

## 2. Problem description and method framework

### 2.1. Robotic batch precision assembly task

This paper focuses on the batch precision assembly task of small-sized, high-precision, non-cylindrical components with equal cross section. Prior to the manufacturing process, tolerance design is carried out for all mating components to ensure the functional requirements of the final assembly. To balance manufacturing cost and machining accuracy, such components are typically designed with a transition fit in the product manufacturing and assembly manual.

A representative batch precision assembly scenario is illustrated in Fig. 1. The dimension of the hexagonal peg, denoted as $d_p$, is defined by the diameter of the circumscribed circle of its cross-section. The dimension of the hexagonal hole is defined as $d_h$ in a similar manner. The nominal values of both $d_p$ and $d_h$ are 10 mm, with a manufacturing tolerance of ±0.02 mm. Following the product manufacturing manual, a large batch of pegs and holes is manufactured. During each assembly process, the robotic system randomly selects a peg and a hole to conduct insertion until all components in the batch have been assembled. In fact, each group of peg and hole exhibits a deterministic but unknown fit amount.

$$f_i \in [d_{p,\min} - d_{h,\max}, d_{p,\max} - d_{h,\min}] \sim \mathcal{N} \tag{1}$$

where $f_i$ is the fit amount of the mating components in the i-th assembly, which approximately follows the normal distribution $\mathcal{N}$ within a certain range. $d_{p,\min}/d_{p,\max}$ and $d_{h,\min}/d_{h,\max}$ denote the lower/upper limits of the peg and hole dimension.

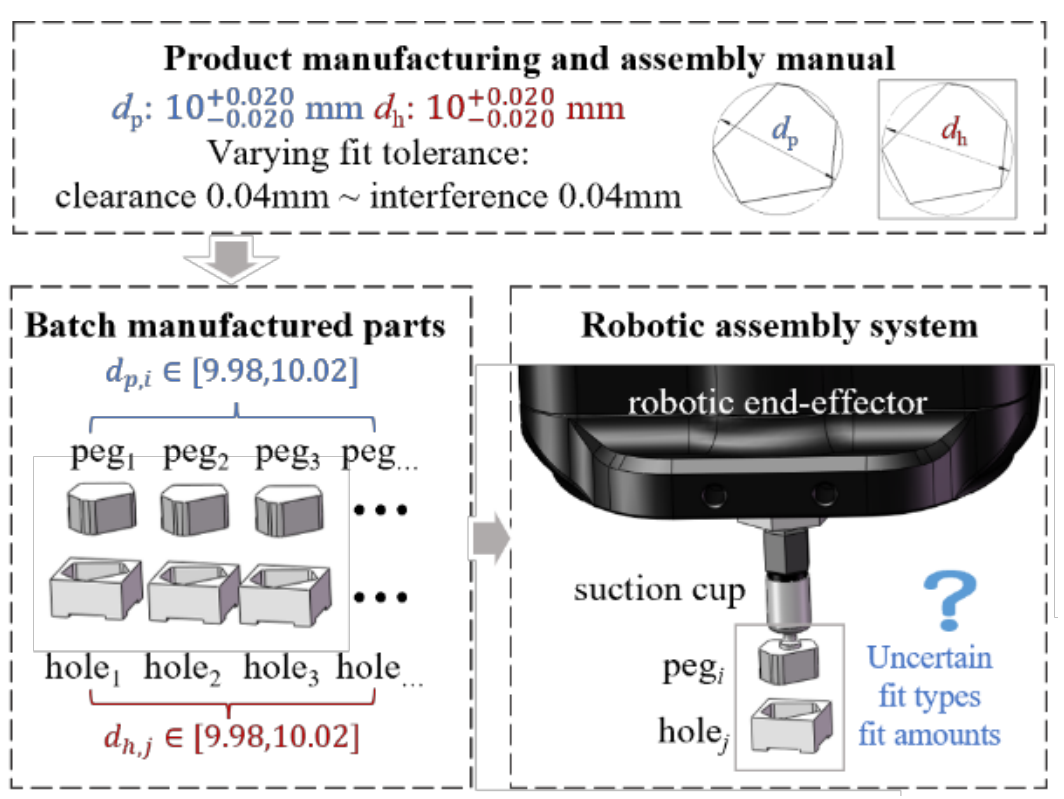

**Fig. 1.** Robotic batch precision assembly task

Different fit amounts essentially lead to significant variations in the environmental dynamics during the assembly process, requiring the robotic assembly control robustness against such uncertainties. In addition, the control strategy should be capable of handling small initial positional and angular misalignments between the peg and hole to achieve successful insertion. As shown in Fig.1, for these high-precision and small-sized parts, the end-effector of the robotic manipulator usually adopts a flexible suction cup instead of a rigid gripper to ensure operability and enhance compliance. Initial alignment errors typically arise from grasping inaccuracies and flexible deformation of the suction cup.

### 2.2. Overall method framework

To construct a control strategy that is robust to the uncertainties in fit amounts and initial misalignments, the most intuitive idea is to directly apply domain randomization to these two parameters in experiments. However, different values of fit amounts result in significantly different contact dynamics, which correspond to substantial variations in environmental transition probabilities in RL, causing challenges to the convergency of the RL training process. In addition, from the perspective of hardware and time cost, real-world domain randomization about the fit amounts requires numerous manufactured parts and the RL training process is extremely slow. Studies in autonomous driving [29] and robotic manipulation [30] have shown that a policy distilled from multiple agents, each trained in a small-scale domain-randomized environment, can achieve better performance and robustness than the policy trained directly in a large-scale domain-randomized environment. Inspired by that, we propose a framework for constructing the robust assembly control strategy, which consists of task decomposition and integration.

As illustrated in Fig. 2, the proposed framework adopts a multi-stage structure. In stage I, several representative parts are generated from the batch precision assembly task to define multiple subtasks. In each subtask, the fit amount between the peg and hole is different and pre-determined, so the domain randomization is applied only to the initial alignment errors. In stage II, RL is employed to learn the optimal control strategy for each subtask. To improve the assembly compliance for high-precision mating components, a force-vision fusion controller (FVFC) is developed as the basic strategy for RL, and the state representation is also augmented to incorporate richer observational visual information. To exploit inter-subtask correlations and improve overall training efficiency, the subtasks are trained simultaneously using a tailored MTRL training framework, instead of independently trained one by one. In stage III, all subtask strategies serve as multiple teachers to distill their knowledges and abilities into a new student network, forming a robust strategy independent of subtask encodings. This distillation process leverages the experience data collected during stage II and is performed via supervised learning, without requiring additional interaction with the assembly environment. Finally, the distillation strategy is independently deployed in the batch precision assembly task and it successfully demonstrates robustness to the uncertain fit types and fit amounts among batch manufactured pegs and holes.

## 3. Method

### 3.1. Task decomposition

As discussed in Section 2.1, while the exact fit amount for each peg-hole pair remains uncertain prior to assembly, its variation range is predetermined in advance, as formulated in Eqs. (1). The strategy learned from a single peg-hole pair with either a clearance fit or an interference fit usually cannot generalize across such a wide range of fit amounts. Therefore, the interval $[d_{p,\min} - d_{h,\max}, d_{p,\max} - d_{h,\min}]$ is divided into multiple smaller sub-intervals, with the intention that the control strategy learned from a representative case in each sub-interval can generalize to the small range of fit amounts contained within the sub-interval. Subsequently, multiple sets of assembly components are generated, with their fit amounts falling within the corresponding sub-intervals, forming multiple deterministic assembly subtasks (labelled as $T_1$, $T_2$, …, $T_n$) for subtask strategy learning. According to the elastic mechanics theory modeling of the assembly process [9], variations in fit amounts induce approximately linear changes in contact forces. Therefore, a linear subdivision scheme is adopted for subintervals to cover the possible force state space uniformly, as formulated in Eqs. (2).

$$\begin{aligned} & f_j \in [d_{p,\min} - d_{h,\max} + (j-1)\kappa, d_{p,\min} - d_{h,\max} + j\kappa] \text{ for } T_j, j = 1,2,\ldots,n \\ & \kappa = \frac{(d_{p,\max} - d_{h,\min}) - (d_{p,\min} - d_{h,\max})}{n} \end{aligned} \tag{2}$$

where $T_j$ denotes the defined $j$-th assembly subtask, $n$ is the total number of subtasks, and $\kappa$ represents the length of the fit amount sub-interval in each subtask. Based on empirical observations, $n$ can be set to 4, such that the divided sub-intervals possess clear physical significance. In this case, the mating components in $T_1$ to $T_4$ are with a large interference fit, a small interference fit, a small clearance fit and a large clearance fit respectively.

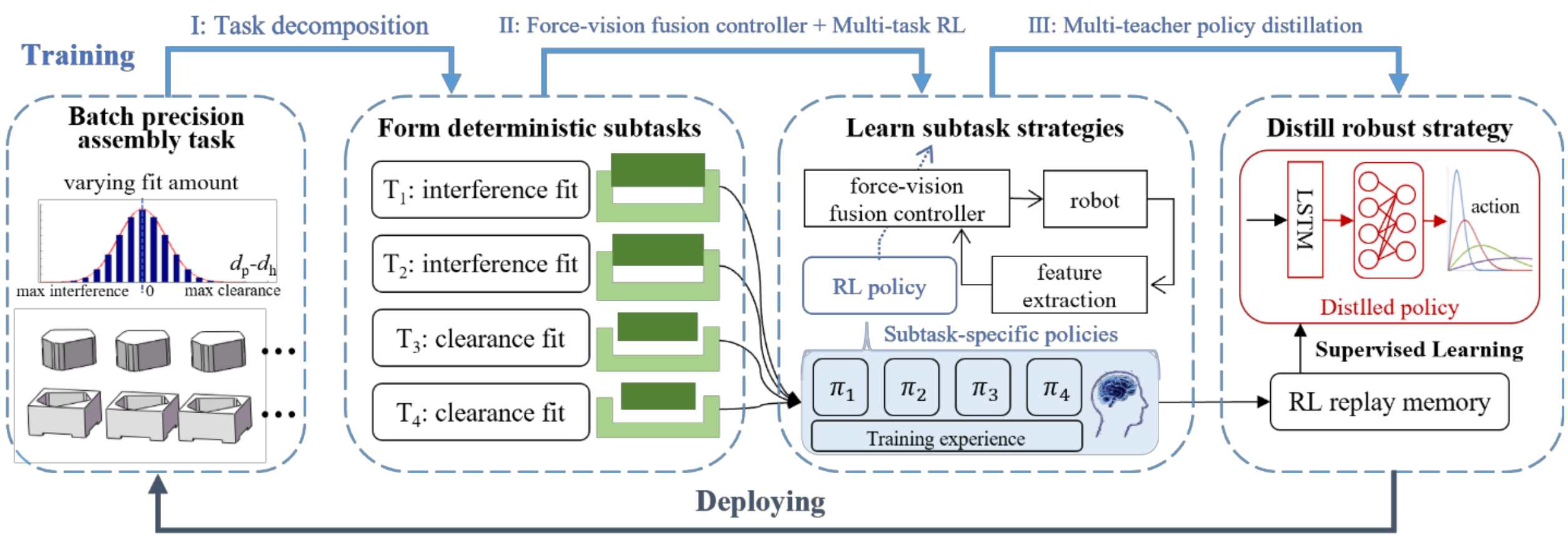

**Fig. 2**. Overall framework to construct the robust assembly control strategy

Direct domain randomization aims to randomly modify the fit amounts of the assembly components before each training episode to produce global robustness, which is hardware-infeasible and time-inefficient. In contrast, task decomposition generates several deterministic subtasks, which are then expected to develop local robustness through multi-task learning.

An addition note is provided here: if the ultimate goal is to construct a control strategy that is robust to different geometrical shapes rather than varying fit amounts, it is only necessary to modify the task decomposition scheme, without requiring any changes to the methods described in Sections 3.2 to 3.4. In this case, each subtask corresponds to an assembly task with a deterministic shape, selected from a pre-defined set of shapes.

### 3.2. Force-vision fusion controller-driven reinforcement learning

Considering each assembly subtask, the objective is to improve the adaptability of the control strategy to random initial misalignments and to enhance the force compliance throughout the assembly process. To this end, force-vision fusion controller-driven reinforcement learning structure is proposed, where the RL agent is trained to adaptively adjust the parameters of a well-defined force-vision fusion controller instead of directly outputting the six-degree-of-freedom (6-D) pose of the robotic manipulator. Furthermore, multimodal perception information—including robot proprioception, force features, and vision features—is incorporated into the RL state representation, thereby improving the perception ability of RL agents. The detailed definitions, modeling, and mathematical formulation are presented in this section.

#### 3.2.1 Assembly system definition

In the assembly scenario illustrated in Fig. 3(a), a six-axis serial robot is used to grasp the peg and assemble it with multiple holes. The assemblies between the single peg and different holes form different subtasks $T_j$.

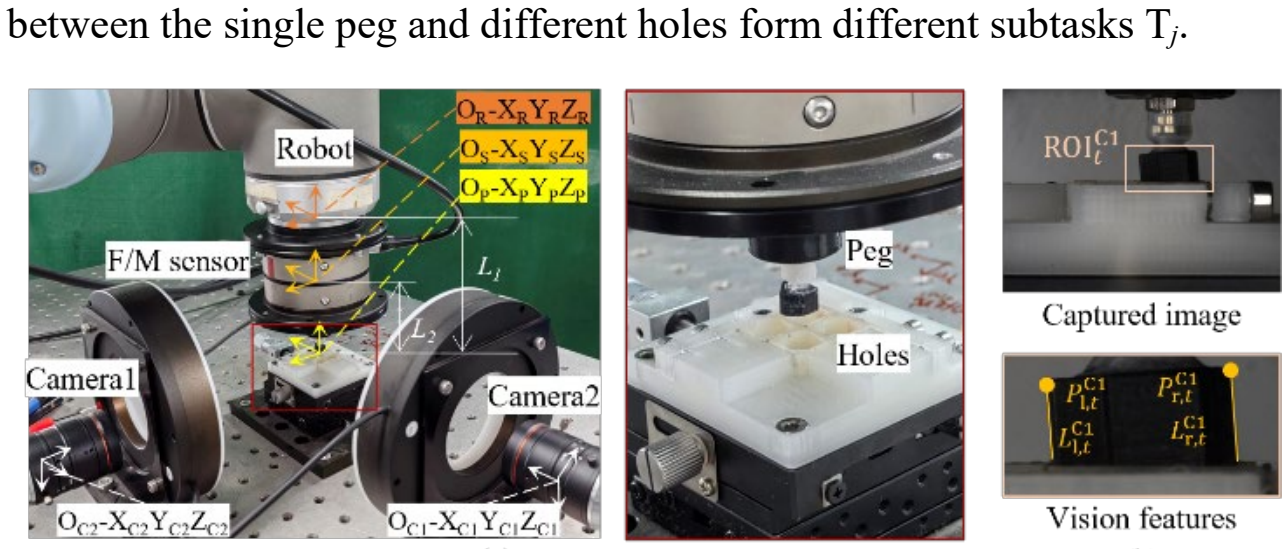

**Fig. 3.** Assembly system and vision features definition

First, the assembly process in each subtask is analyzed independently. As shown in Fig. 3(a), several coordinate systems are defined to assist in describing the robotic assembly process. The robot coordinate system $O_R$-$X_RY_RZ_R$ is established to control the movement of the robot manipulator. The coordinate system $O_S$-$X_SY_SZ_S$ is defined at the measurement center of the force/moment (F/M) sensor to reflect forces and moments during the assembly process. Two orthogonal monocular cameras are employed to capture images, with their coordinate systems denoted as $O_{C1}$-$X_{C1}Y_{C1}Z_{C1}$ and $O_{C2}$-$X_{C2}Y_{C2}Z_{C2}$, respectively. The $O_P$-$X_PY_PZ_P$ is a dynamic coordinate system, which is initially located at the center of the peg's bottom surface before assembly. During the assembly process, $O_P$-$X_PY_PZ_P$ is adjusted to the center of the peg's insertion region into the hole. The $O_P$-$X_PY_PZ_P$ is designed to estimate the actual contact moments and to guide pose adjustments more accurately. There are vertical offsets along the Z-axis between $O_P$, $O_S$ and $O_R$ before assembly, denoted by $L_1$ and $L_2$.

#### 3.2.2 Force-vision fusion controller design

The control block diagram of the FVFC is shown in Fig. 4. The control objective is to maintain small contact forces and pose errors throughout the continuous insertion process. The control command of the FVFC is the Cartesian pose of the coordinate frame $O_R$-$X_RY_RZ_R$ relative to the robot base frame, denoted as $p_R(t)$.

$$p_R(t) = [x_{R,t}, y_{R,t}, z_{R,t}, \alpha_{R,t}, \beta_{R,t}, \gamma_{R,t}]^T \tag{3}$$

where $x_{R,t}$, $y_{R,t}$, $z_{R,t}$ are the translation control components, and $\alpha_{R,t}$, $\beta_{R,t}$, $\gamma_{R,t}$ are the RPY (Roll, Pitch, Yaw) angle control components.

A flexible suction cup is used as the gripper between the robot and the peg. In each control cycle, the robot indirectly controls the peg and the hole to interact with each other. Key features are extracted via force and vision sensors, and subsequently fed into the FVFC for the computation of $p_R(t)$.

The 6-D force and moment signal $F_S(t)$ is directly measured by the F/M sensor. Mean filtering is applied to suppress random noise. After gravity compensation, the actual contact force and moment $F_S'(t)$ can be calculated.

$$F_S'(t) = [F_{x,t}, F_{y,t}, F_{z,t}, M_{x,t}, M_{y,t}, M_{z,t}]^T \tag{4}$$

where $F_{x,t}$, $F_{y,t}$, $F_{z,t}$ are the force components and $M_{x,t}$, $M_{y,t}$, $M_{z,t}$ are the moment components of $F_S'(t)$.

It is noted that $F_S'(t)$ is in the sensor frame $O_S$-$X_SY_SZ_S$. To more accurately reflect the pose error between the peg and the hole, a coordinate transformation module is designed to estimate the equivalent force and moment $F_{\mathrm{P}}(t)$ in the frame $O_P$-$X_PY_PZ_P$. $L_2$ is always parallel to $Z_S$, so its force arm is accurate. This transformation module is approximate because the initial alignment angle error between the peg and hole exists, and the suction cup may undergo angular deformation during assembly. However, since the angular error is relatively small (less than two degrees) and $\hat{l}_t \ll L_2$, the resulting moment arm error on $\hat{l}_t$ is considered negligible.

$$F_{\mathrm{P}}(t) \approx [F_{x,t}, F_{y,t}, F_{z,t}, M_{x,t} - F_{y,t}(L_2 - \frac{\hat{l}_t}{2}), M_{y,t} + F_{x,t}(L_2 - \frac{\hat{l}_t}{2}), M_{z,t}]^T \tag{5}$$

where $\hat{l}_t$ denotes the insertion depth of the peg into the hole, which can be estimated based on the movement of the robot along the $Z_R$ direction.

$$\hat{l}_t = \max(0, z_{\mathrm{R},0} - z_{\mathrm{R},t} - h_0) \tag{6}$$

where $h_0$ denotes the initial height from the bottom surface of the peg to the upper surface of the hole before assembly, typically a small value.

Since $F_{\mathrm{P}}(t)$ can more effectively reflect the contact state between the peg and hole, a proportional controller is adopted as the force controller.

$$\begin{aligned} &\Delta p_{\mathrm{F}}(t) = \mathbf{K}_{\mathrm{F}}(F_{\mathrm{P}}(t) - F_{\mathrm{ref}}) + [0, 0, \Delta z_{\mathrm{ref}}, 0, 0, 0]^T \\ &\mathbf{K}_{\mathrm{F}} = \mathrm{diag}(k_{\mathrm{F},x}, k_{\mathrm{F},y}, k_{\mathrm{F},z}, k_{\mathrm{M},x}, k_{\mathrm{M},x}, k_{\mathrm{M},x}) \end{aligned} \tag{7}$$

where $\mathbf{K}_{\mathrm{F}}$ is the 6×6 diagonal stiffness matrix of the force controller, $\Delta z_{\mathrm{ref}}$ is the pre-defined reference insertion depth at each control step, and $F_{\mathrm{ref}}$ is the reference insertion force.

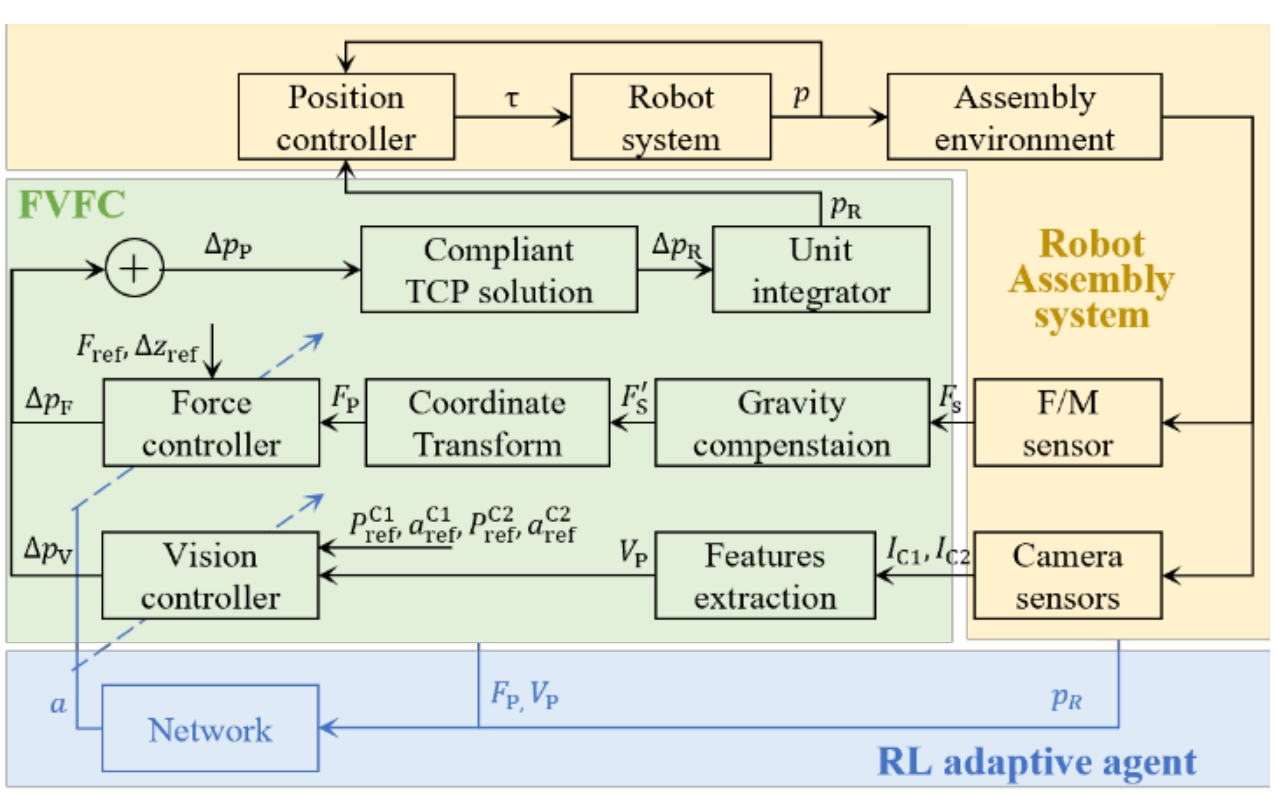

**Fig. 4.** Control block diagram of FVFC-driven RL strategy

On the other hand, based on two orthogonally arranged monocular cameras, visual features are extracted to characterize the true relative pose between the peg and hole under suction cup deformation conditions. At each control step, two images, $I_{\mathrm{C1}}(t)$ and $I_{\mathrm{C2}}(t)$, are captured, and the vision features $V_{\mathrm{P}}(t)$ are computed using a feature extraction module. As illustrated in Fig. 3(b), the feature extraction module consists of a series of fast traditional computer vision algorithms, including time-varying region of interest (ROI) extraction, adaptive binary segmentation and Ed-Lines detection [31]. For $I_{\mathrm{C1}}(t)$, two corner points ( $P_{\mathrm{l},t}^{\mathrm{C1}}$, $P_{\mathrm{r},t}^{\mathrm{C1}}$ ) and two edge lines ( $L_{\mathrm{l},t}^{\mathrm{C1}}$, $L_{\mathrm{r},t}^{\mathrm{C1}}$ ) of the peg are finally accurately extracted. For $I_{\mathrm{C2}}(t)$, $P_{\mathrm{l},t}^{\mathrm{C2}}$, $P_{\mathrm{r},t}^{\mathrm{C2}}$, $L_{\mathrm{l},t}^{\mathrm{C2}}$, $L_{\mathrm{r},t}^{\mathrm{C2}}$ are defined and acquired using the same procedure. Then the complete visual feature vector $V_{\mathrm{P}}(t)$ is calculated by the following equation.

$$\begin{aligned} V_{\mathrm{P}}(t) = \frac{1}{2}[&\left(P_{\mathrm{l},t}^{\mathrm{C2}}(u) + P_{\mathrm{r},t}^{\mathrm{C2}}(u) - 2P_{\mathrm{ref}}^{\mathrm{C2}}(u)\right), \left(P_{\mathrm{l},t}^{\mathrm{C1}}(u) + P_{\mathrm{r},t}^{\mathrm{C1}}(u) - 2P_{\mathrm{ref}}^{\mathrm{C1}}(u)\right), \\ &\left(\mathrm{A}(L_{\mathrm{l},t}^{\mathrm{C1}}) + \mathrm{A}(L_{\mathrm{r},t}^{\mathrm{C1}}) - 2a_{\mathrm{ref}}^{\mathrm{C1}}\right), -\left(\mathrm{A}(L_{\mathrm{l},t}^{\mathrm{C2}}) + \mathrm{A}(L_{\mathrm{r},t}^{\mathrm{C2}}) - 2a_{\mathrm{ref}}^{\mathrm{C2}}\right)]^T \end{aligned} \tag{8}$$

where $P(u)$ represents the horizontal coordinate of an image point $P$, and A() is the function to calculate the inclination angle of an image line. $P_{\mathrm{ref}}^{\mathrm{C1}}$, $P_{\mathrm{ref}}^{\mathrm{C2}}$, $a_{\mathrm{ref}}^{\mathrm{C1}}$ and $a_{\mathrm{ref}}^{\mathrm{C2}}$ reflect the observed pose information of the hole in the image, which remain unchanged during the assembly process. In industrial scenario, holes are typically fixed by mechanical fixtures, so the corresponding hole parameters can be pre-calibrated in advance to improve the computational efficiency of the feature extraction algorithm.

Essentially, $V_{\mathrm{P}}(t)$ can accurately represent the translation and rotation error between the peg and hole even when the suction cup is deformed. Moreover, it can implicitly indicate the occurrence of jamming during the assembly process. Based on $V_{\mathrm{P}}(t)$, the vision controller produces additional translational and rotational adjustments along the $X_P$ and $Y_P$ direction. It enhances the system's adaptability to initial alignment errors and compensate for the unobservability of the relative pose induced when suction cup deformation.

$$\Delta p_{\mathrm{V}}(t) = \begin{bmatrix} k_{\mathrm{V},x} & 0 & 0 & 0 & 0 & 0 \\ 0 & k_{\mathrm{V},y} & 0 & 0 & 0 & 0 \\ 0 & 0 & 0 & k_{\mathrm{V},\alpha} & 0 & 0 \\ 0 & 0 & 0 & 0 & k_{\mathrm{V},\beta} & 0 \end{bmatrix}^T \begin{bmatrix} k_{\mathrm{cb}}^{\mathrm{C2}} & 0 & 0 & 0 \\ 0 & k_{\mathrm{cb}}^{\mathrm{C1}} & 0 & 0 \\ 0 & 0 & 1 & 0 \\ 0 & 0 & 0 & 1 \end{bmatrix} V_{\mathrm{P}}(t) \tag{9}$$

where $k_{\mathrm{V},x}$, $k_{\mathrm{V},y}$, $k_{\mathrm{V},\alpha}$ and $k_{\mathrm{V},\beta}$ are the parameters of the vision controller. $k_{\mathrm{cb}}^{\mathrm{C1}}$ and $k_{\mathrm{cb}}^{\mathrm{C2}}$ are pre-calibrated coefficients that reflect scale correspondence from the pixel coordinate system to the frame $O_P$-$X_PY_PZ_P$.

The expected relative motion of the peg $\Delta p_{\mathrm{P}}(t)$ is the element-wise addition of the outputs from the force controller and the vision controller. Subsequently, the corresponding motion command of the robot end-effector $\Delta p_{\mathrm{R}}(t)$ is computed through coordinate transformation.

$$\Delta p_{\mathrm{P}}(t) = \Delta p_{\mathrm{F}}(t) + \Delta p_{\mathrm{V}}(t) = [\Delta x_{\mathrm{P},t}, \Delta y_{\mathrm{P},t}, \Delta z_{\mathrm{P},t}, \Delta \alpha_{\mathrm{P},t}, \Delta \beta_{\mathrm{P},t}, \Delta \gamma_{\mathrm{P},t}]^T \tag{10}$$

$$\begin{aligned} \Delta p_{\mathrm{R}}(t) = [&\Delta x_{\mathrm{P},t} + (L_1 - \frac{\hat{l}_t}{2})\sin(\Delta \beta_{\mathrm{P},t}), \Delta y_{\mathrm{P},t} - (L_1 - \frac{\hat{l}_t}{2})\sin(\Delta \alpha_{\mathrm{P},t}), \\ &\Delta z_{\mathrm{P},t}, \Delta \alpha_{\mathrm{P},t}, \Delta \beta_{\mathrm{P},t}, \Delta \gamma_{\mathrm{P},t}]^T \end{aligned} \tag{11}$$

Finally, the absolute pose control command $p_{\mathrm{R}}(t)$ is obtained through a unit integrator.

### 3.2.3 Combining RL with FVFC

Compared to the force controller, FVFC comprehensively leverages both vision and force features to obtain higher control performance. However, the controller parameters tuning remains a critical issue. Due to the time-varying and nonlinear assembly dynamics, manually pre-set constant parameters usually cannot achieve the optimal control performance—especially for those components with irregular cross-section shapes. To address this issue, a RL agent is combined with the FVFC, enabling adaptive adjustment of the controller parameters to improve the comprehensive control performance.

The state of the Markov decision process (MDP) is defined by the robot's relative pose, force and vision information. When contact occurs between the peg and the hole, the flexible suction cup will undergo uncontrollable soft deformation. Additional vision feature $V_{\mathrm{P}}(t)$ can help enrich the observation of the actual relative pose between the peg and hole. The action in the MDP is defined as the adjustment of the FVFC controller parameters.

$$\overline{s_t} = [p_{\mathrm{R}}(t) - p_{\mathrm{R}}(0), F_{\mathrm{P}}(t), V_{\mathrm{P}}(t)] \tag{12}$$

$$a_t = [\Delta k_{\mathrm{F},x}, \Delta k_{\mathrm{F},y}, \Delta k_{\mathrm{F},z}, \Delta k_{\mathrm{M},x}, \Delta k_{\mathrm{M},x}, \Delta k_{\mathrm{M},x}, \Delta k_{\mathrm{V},x}, \Delta k_{\mathrm{V},y}, \Delta k_{\mathrm{V},\alpha}, \Delta k_{\mathrm{V},\beta}] \tag{13}$$

The actual controller parameters in Eqs. (7) and (9) become time-varying, equal to the sum of the manually pre-set parameters $\hat{k}$ and the adaptive values $a_t$ output by the RL agent.

$$k_t = \hat{k} + a_t \tag{14}$$

where $k_t = [k_{F,x}, k_{F,y}, k_{F,z}, k_{M,x}, k_{M,x}, k_{M,x}, k_{V,x}, k_{V,y}, k_{V,\alpha}, k_{V,\beta}]_t$ is the actual controller parameters used for FVFC at each control step.

The reward function for RL is defined as follows.

$$r_t = r_i + r_{FM} + r_{stp} + r_s$$
$$\begin{cases} r_i = k_i \Delta z_{R,t} \\ r_{FM} = -k_F \|[F_{x,t}, F_{y,t}, F_{z,t}]\|_1 - k_M \|[M_{x,t}, M_{y,t} M_{z,t}]\|_1 \\ r_{stp} = -k_{stp} \\ r_S = \begin{cases} k_s, \text{if success} \\ k_f, \text{if failure} \end{cases} \end{cases} \tag{15}$$

Where $r_i$, $r_{FM}$, $r_{stp}$, $r_S$ denote the reward for insertion, the penalty for force and moment, the penalty for the insertion steps, and the sparse reward given upon success or failure at the end of each episode, respectively. The corresponding reward coefficients are manually set. In subtasks where the peg and hole are in an interference fit, $r_{FM}$ does not involve the penalty on the axial force $F_{z,t}$, as a large axial force is inherent and avoidable in interference assembly subtasks.

During the assembly process, the failure occurs if $|F_S^{'}(t)|$, $|V_P(t)|$ or total number of control steps exceeds the predefined threshold $F_{S,\max}^{'}$, $V_{P,\max}$, and $T_{\max}$. Conversely, the assembly is considered successful if the insertion depth $H$ is successfully reached.

### 3.3. Efficient MTRL training method

The FVFC-driven RL can be applied to each subtask $T_j$ to independently learn a compliance control strategy. However, single-task reinforcement learning (STRL) is time-inefficient. Considering the inherent correlations among different assembly subtasks, a MTRL training method is developed to implicitly learn shared feature representations, thereby improving overall training efficiency.

The MTRL training architecture is illustrated in Fig. 5(a). Multi-task soft actor-critic (MTSAC) [32] is chosen as the basic algorithm. The actor and critic network are shared across all subtasks. Disentangled entropy coefficients are assigned to each subtask to support task-specific exploration. A two-layer Long Short-Term Memory (LSTM) network is adopted as the feature extractor in both the actor and critic networks to extract the temporal information during the assembly process. The temporal state $s_t$ is the multi-step stack of the original state $\bar{s}_t$ defined in Eqs. (12). To distinguish different subtasks, subtask encodings are added to the collected state $s_t$, forming the augmented state $\tilde{s}_t$, which serves as the actual input to both the actor and critic network.

$$\begin{aligned} s_t &= [\bar{s}_{t-m+1}, \bar{s}_{t-m+2}, \ldots, \bar{s}_t] \\ \tilde{s}_t &= TC_j(s_t) = [\bar{s}_{t-m+1}, c_{T_j}, \bar{s}_{t-m+2}, c_{T_j}, \ldots, \bar{s}_t, c_{T_j}], \text{ if } s_t \text{ is from } T_j \end{aligned} \tag{16}$$

where $m$ is the number of stack steps, $c_{T_j}$ is the one-hot encoding of task $T_j$, and $TC_j()$ is defined as the task encoder function for task $T_j$.

$$c_{T_j} = [\underbrace{0, \ldots, 0}_{j-1}, 1, \underbrace{0, \ldots, 0}_{n-j}] \tag{17}$$

In terms of transition collection, separate replay buffers ($RB_1$~$RB_n$) are assigned to different subtasks. Since only a single robot is used to perform different assembly subtasks in the experiment scenario, transitions must be collected sequentially—i.e., a new subtask can be executed only after the current one completes at least an episode. Rather than executing subtasks in a predetermined order, a task selector is designed to dynamically choose the next subtask， thereby enabling more balanced sampling across all subtasks.

$$p(T_j) = \frac{N(RB_j)}{\sum_{i=1}^{n} N(RB_i)}, j = 1, 2, \ldots, n \tag{18}$$

where $p(T_j)$ is the probability that the task selector chooses $T_j$, and $N(RB_i)$ is the number of transitions in $RB_i$.

At each timestep, a transition $[s_t, a_t, r_t, s_{t+1}]$ is collected from the current subtask and processed by the task encoding module to obtain the encoded transition $[\tilde{s}_t, a_t, r_t, \tilde{s}_{t+1}]$, which is then stored in the corresponding replay buffer.

During the training process, a mini-batch is sampled from each subtask's replay buffer, denoted as $B_1$, $B_2$, …, $B_n$. For each subtask $T_j$, the gradient of the actor network is computed as $\widehat{\nabla}_\phi\left(\mathcal{L}_{\text{actor}}\right)_{\mathrm{T}_j}$ , and the gradient of the critic network is computed as $\widehat{\nabla}_{\theta_i}\left(\mathcal{L}_{\text{critic}}\right)_{\mathrm{T}_j}, i=1,2$ . The relevant loss functions are as follows, and the gradients are easily obtained via automatic differentiation using Pytorch.

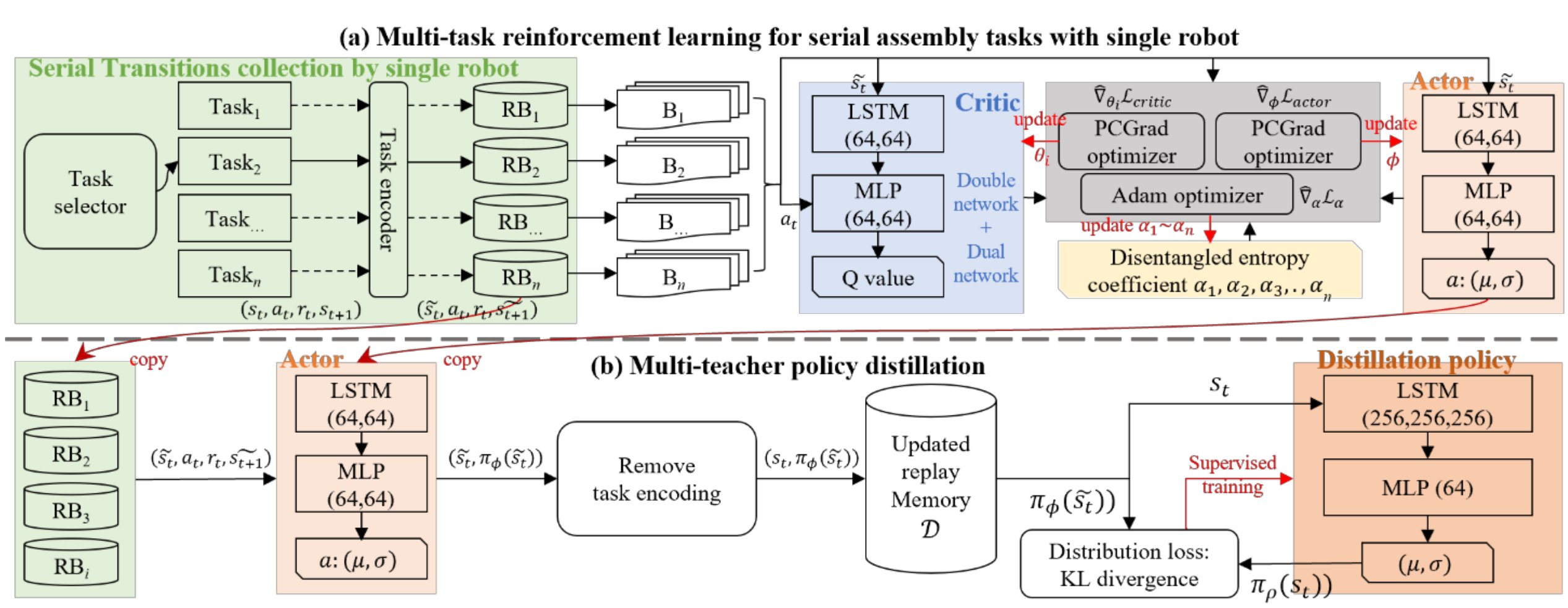

**Fig. 5.** The implementation of MTRL training and policy distillation.

$$
\begin{aligned}
&\left(\mathcal{L}_{\text{actor}}\right)_{\mathrm{T}_j}=E_{\tilde{s}_t\sim B_j}\left[E_{a\sim\pi_\phi(\cdot|\tilde{s}_t)}\left(\alpha_j\log\pi_\phi(a\mid\tilde{s}_t)-\min_{i=1,2}Q_{\theta_i}(\tilde{s}_t,a)\right)\right]\\
&\left(\mathcal{L}_{\text{critic}}\right)_{\mathrm{T}_j}=\frac{1}{2}\sum_{i=1}^{2}E_{(\tilde{s}_t,a_t,r_t,\tilde{s}_{t+1})\sim B_j}\left[\left(Q_{\theta_i}(\tilde{s}_t,a_t)-E_{a\sim\pi_\phi(\cdot|\tilde{s}_{t+1})}Q_{\text{target}}\right)^2\right]\\
&\quad Q_{\text{target}}=r_t+\gamma\left[\min_{i=1,2}Q^{\text{target}}_{\bar{\theta}_i}(\tilde{s}_{t+1},a)-\alpha_j\log\pi_\phi(a\mid\tilde{s}_{t+1})\right]
\end{aligned}
\tag{19}
$$

where $Q_{\theta_1}, Q_{\theta_2}$ are the online Q networks, and $Q^{\text{target}}_{\bar{\theta}_1}, Q^{\text{target}}_{\bar{\theta}_2}$ are the target networks. $\pi_\phi$ is the actor network, and $a\sim\pi_\phi(\cdot\mid\tilde{s}_t)$ means sampling an action from a spherical Gaussian distribution with mean and standard deviation given by $\pi_\phi(\tilde{s}_t)$ . $\alpha_j$ is the task-specific entropy coefficient for $\mathrm{T}_j$ , and $\gamma$ is the discount factor.

Then Projecting Conflicting Gradients (PCGrad) [33] is employed to mitigate gradient conflicts among different subtasks to stabilize training process. The essence of PCGrad is to identify and adjust the conflicting items between the gradients computed from different subtasks.

$$
\begin{aligned}
&\widehat{\nabla}_\phi\mathcal{L}_{\text{actor}}=\text{PCGrad}\left[\widehat{\nabla}_\phi\left(\mathcal{L}_{\text{actor}}\right)_{T_j}, j=1,2,...,n\right]\\
&\widehat{\nabla}_{\theta_i}\mathcal{L}_{\text{critic}}=\text{PCGrad}\left[\widehat{\nabla}_{\theta_i}\left(\mathcal{L}_{\text{critic}}\right)_{T_j}, j=1,2,...,n\right], i=1,2
\end{aligned}
\tag{20}
$$

where $\widehat{\nabla}_\phi\mathcal{L}_{\text{actor}}$ and $\widehat{\nabla}_{\theta_i}\mathcal{L}_{\text{critic}}$ are the final gradients of the actor and critic networks, which are used for updating the network parameters.

The disentangled entropy coefficients are automatically adjustment, and the gradient $\nabla\mathcal{L}_\alpha$ is calculated as follows:

$$
\begin{aligned}
&\left(\mathcal{L}_{\alpha_j}\right)_{T_j}=E_{\tilde{s}_t\sim B_j}\left[E_{a\sim\pi_\phi(\cdot|\tilde{s}_t)}\left(-\alpha_j\log\pi_\phi(a\mid\tilde{s}_t)-\alpha_j\overline{\mathcal{H}}\right)\right]\\
&\nabla\mathcal{L}_\alpha=\nabla\left[\sum_{j=1}^{n}\left(\mathcal{L}_{\alpha_j}\right)_{T_j}\right], \alpha=[\alpha_1,\alpha_2,...,\alpha_n]
\end{aligned}
\tag{21}
$$

where $\overline{\mathcal{H}}$ is the desired maximum expected entropy.

In terms of update frequency, whenever a new transition is collected from an arbitrary subtask, all replay buffers are utilized to perform one update of the actor network, the critic network, and the entropy coefficients by Eqs. (19)-(21).

By the MTRL training method, a single policy network is trained, which can capture the implicit correlation among all subtasks. However, since the input state $\tilde{s}_t$ of the policy network contains the subtask encoding, the learned policy can be used only for the given subtasks and cannot be directly generalized to unseen ones. Essentially, multiple strategies for individual subtasks are learned, which can be represented as follows.

$$
\pi_j(s_t)=\pi_\phi(TC_j(s_t)), j=1,2,....,n \tag{22}
$$

where $\pi_j$ is the trained compliant control strategy for subtask $\mathrm{T}_j$ .

### 3.4. Policy distillation for robustness strategy

In this section, the robust control strategy for assembly with uncertain fit amount is constructed through policy distillation. Specifically, the already learned policies, $\pi_j(s_t), j=1,2,....,n$ , serve as multiple teacher models, which collectively transfer their knowledge into a new student network through supervised learning.

As shown in Fig. 5(b), the replay buffers of all subtasks during MTRL training are saved and reused. For each transition $[\tilde{s}_t,a_t,r_t,\tilde{s}_{t+1}]$ from the replay buffers, the trained MTRL actor network is utilized to generate a new optimal action $\pi_\phi(\tilde{s}_t)$ for $\tilde{s}_t$ . To obtain the robustness of the final policy, the task encodings in $\tilde{s}_t$ are removed to recover $s_t$ .

$$
\left(s_t=TC_j^{-1}(\tilde{s}_t),\ \text{for}\ \tilde{s}_t\sim RB_j\right), j=1,2,...,n \tag{23}
$$

By processing all data from the replay buffers according to the above steps, a labeled dataset $\mathcal{D}:(s_t,\pi_\phi(\tilde{s}_t))$ is obtained. Subsequently, $\mathcal{D}$ is used for supervise training of a new policy network $\pi_\rho$ , whose outputs are the mean and standard deviation of the action Gaussian distribution. The network $\pi_\rho$ adopts a deeper and wider LSTM module, aiming to implicitly infer the fit amount from the current assembly trajectory of vision and force. The training loss function is defined by the Kullback–Leibler (KL) divergence.

$$
\mathcal{L}_{\pi_\rho}=\text{KL}\left(\pi_\phi(\tilde{s}_t)\,\|\,\pi_\rho(s_t)\right) \tag{24}
$$

Theoretically, only reusing data from replay memory is a suboptimal distillation approach, as the state distribution may differ from that encountered during actual deployment. Involving additional real-world data collection using the distillation policy can address the state distribution bias, but brings numerous physical interactions. Subsequent experiments demonstrate that the former approach achieves satisfactory performance, thus obviating the necessity for supplementary sampling.

Finally, the distillation network can be directly applied to any assembly subtask and is capable of generalizing to batch precision assembly task with uncertain fit amounts in the varying range presented in Eqs. (1). Essentially, the distillation network integrates the diverse state spaces of multiple subtasks and the capabilities of all the learned sub-strategies.

## 4. Real-world experiments

In this section, a batch precision assembly task of irregular hexagonal components with uncertain fit types and fit amounts is adopted to validate the effectiveness of the proposed method. The robust control strategy is constructed to handle uncertain fit types and fit amounts, and it is validated that the resulting control strategy can further generalize to uncertain fit amounts.

The proposed method (abbreviated as FVFC-MTRL-PD) is comprehensively compared with several baseline approaches: FBCC [9], a constant-parameter compliance force control algorithm for high-precision assembly task; MDRL [16], a model-driven reinforcement learning control algorithm which combines STRL with a well-designed force controller; and FVFC-CP, the constant-parameter FVFC proposed in section 3.2. The comparison primarily focuses on the robustness and compliance of the control strategies, as well as the efficiency of the RL training process.

### 4.1. Experimental system setup

The experimental system is illustrated in Fig. 6 (a). A UR10 robot is used as the manipulator, equipped with an SRI M3815B F/M sensor. An SMC suction cup is attached to the end-effector of the robot, and a peg is glued to the suction cup for RL training. The vision system comprises two monocular cameras (Hik MV-CH120-10UC), paired with two lenses (Hik MVL-KF5028M-12MPE) and two white light sources. The computer configuration is as follows: Ubuntu 20.04 system, 13th Gen Intel Core i7-13700 CPU, and NVIDIA GeForce RTX 4080 GPU.

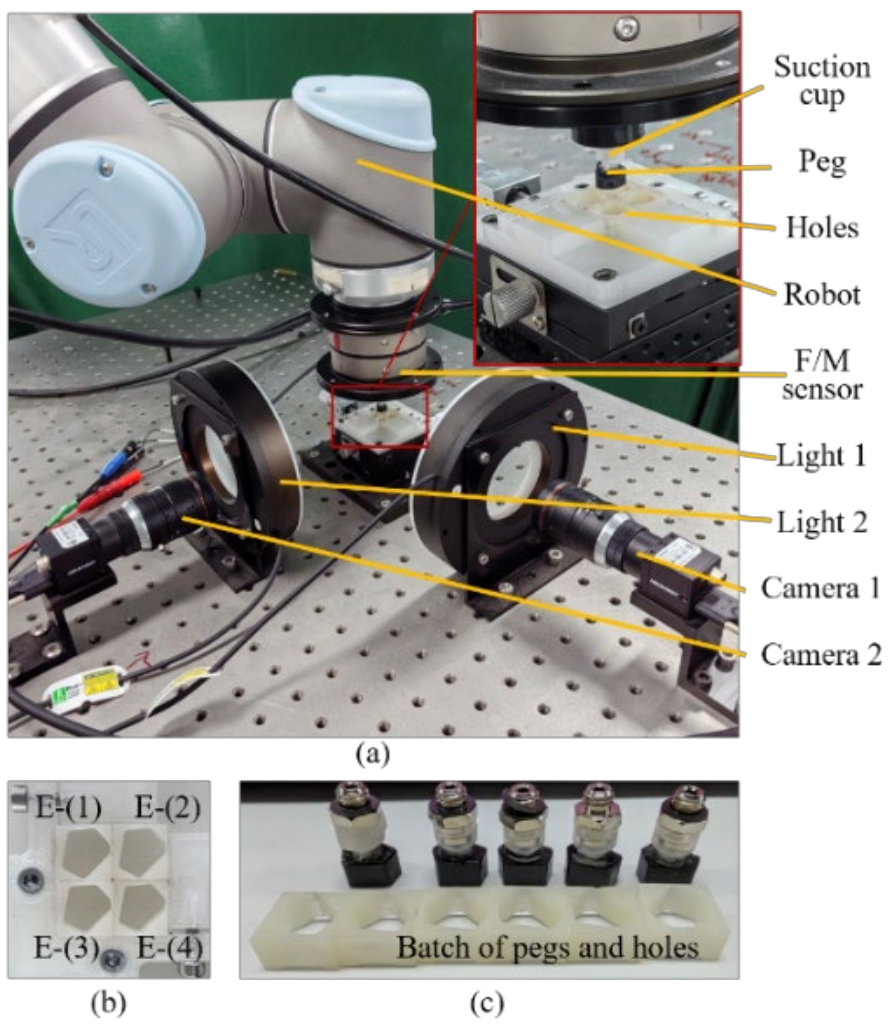

**Fig. 6** Experiment setup and assembly components.

Four holes with different dimensions are utilized for robust control strategy construction, as shown in Fig. 6 (b). These four holes, combined with the peg affixed to the suction cup, define four assembly subtasks ($T_1$ ~ $T_4$) with different fit types and fit amounts, which are detailed in Table 1. The cross-section shape of both the peg and holes is an irregular hexagon. The peg height and hole depth are both 5mm, with nominal cross-sectional dimensions of 10 mm. There is a 0.1 mm chamfer on the edge of each hole. Essential experimental parameters are summarized in Table 2.

The final constructed control strategy will be further applied to the batch-manufactured components shown in Fig. 6 (c) to evaluate the robustness and compliance. The fit amounts of these components are unknown but lie within the range of -0.04 mm to 0.04 mm.

**Table 1 Fit types and fit amounts in four subtasks in experiment.**

| Subtask id | Component id | Fit types | $f$/mm |
|---|---|---|---|
| $T_1$ | E-(1) | interference | -0.04 ~ -0.02 |
| $T_2$ | E-(2) | interference | -0.02 ~ 0.00 |
| $T_3$ | E-(3) | clearance | 0.00 ~ 0.02 |
| $T_4$ | E-(4) | clearance | 0.02 ~ 0.04 |

**Table 2 Parameters in real-world experiment.**

| Parameters | Value |
|---|---|
| FVFC control parameters | $\Delta z_{\mathrm{ref}} = -5e^{-5}$ m; $F_{\mathrm{ref}} = [0,0,0,0,0,0]$ N<br>$k_{\mathrm{F},x}, k_{\mathrm{F},y} \in [-1e^{-5}, 6e^{-5}]$ m/N, $k_{\mathrm{F},z} \in [0, 3e^{-5}]$ m/N<br>$k_{\mathrm{M},x}, k_{\mathrm{M},y}, k_{\mathrm{M},z} \in [-3.5e^{-2}, 3.5e^{-2}]$ rad/Nm<br>$k_{\mathrm{V},x}, k_{\mathrm{V},y}, \in [-1e^{-2}, 6e^{-2}]$<br>$k_{\mathrm{V},\alpha}, k_{\mathrm{V},\beta} \in [-1e^{-2}, 3e^{-2}]$ |
| State stack steps | $m = 5$ |
| Reward coefficients | $k_{\mathrm{i}} = 2e^{-3}\ \mathrm{m}^{-1}$, $k_{\mathrm{F}} = 0.2\ \mathrm{N}^{-1}$, $k_{\mathrm{M}} = 2.0\ \mathrm{Nm}^{-1}$<br>$k_{\mathrm{stp}} = 0.05$, $k_{\mathrm{s}} = +20$, $k_{\mathrm{f}} = -10$ |
| Assembly failure threshold | $F'_{\mathrm{S,max}} = [2\ \mathrm{N}, 2\ \mathrm{N}, 5\ \mathrm{N}, 0.2\ \mathrm{Nm}, 0.2\ \mathrm{Nm}, 0.2\ \mathrm{Nm}]$<br>$V_{\mathrm{P,max}} = [60\ \mathrm{pixel}, 60\ \mathrm{pixel}, 5\ °, 5\ °]$, $T_{\mathrm{max}} = 150$ |
| Assembly depth | $H = 4.0$ mm |
| Random pose range | ±0.3 mm / ±0.5° ($T_1$~$T_3$); ±0.4 mm / ±0.5° ($T_4$) |
| MTRL learning rate | 3e-4 |
| Learning batch size | 256 |

### 4.2. Training process

During the construction process of the control strategy, FVFC-driven RL and MTRL training are first jointly applied to the four subtasks with different fit types and fit amounts to efficiently obtain optimal control strategies for each. These four strategies, referred to as FVFC-MTRL, can be independently deployed to the corresponding subtasks given the task encodings, but they lack robustness across all subtasks. Subsequently, policy distillation is used to derive the final robust control strategy, termed FVFC-MTRL-PD. It eliminates the dependency on task encoding and can be directly applied to the batch precision assembly task.

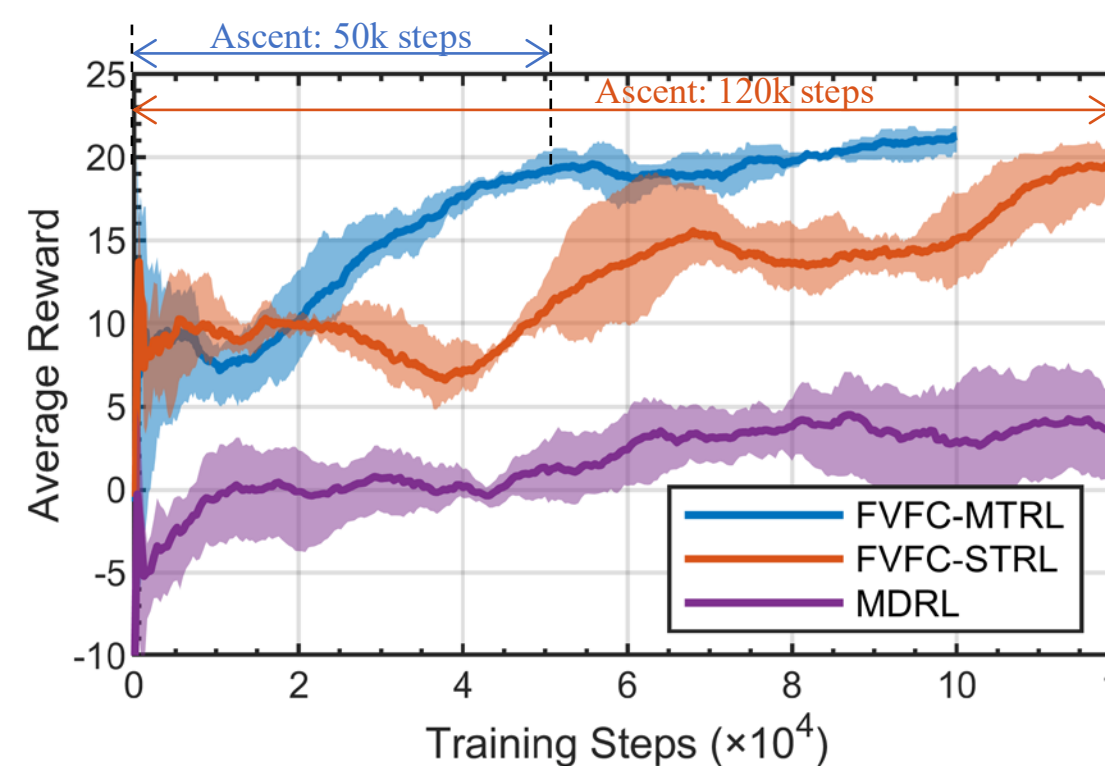

**Fig. 7** Average reward for four subtasks during training process

The training process of FVFC-MTRL is first analyzed and compared with FVFC-STRL (which trains four subtasks individually using FVFC-driven RL), and MDRL. The comparison is conducted to evaluate both the performance and the training efficiency. The training process of each RL algorithm is repeated three times using different random seeds. For each RL algorithm, the mean reward trajectories are plotted as solid curves and the area between the minimum and maximum values is filled with a color block, as illustrated in Fig. 7. Compared to the FVFC-STRL/FVFC-MTRL method, the final reward of the MDRL is significantly lower, which essentially demonstrates the superior assembly ability of FVFC compared with force-based controlled. On the other hand, FVFC-MTRL achieves obviously faster training convergence and greater stability compared to FVFC-STRL. The average reward across four subtasks in FVFC-MTRL increases steadily and converges at approximately 50k steps. However, the reward of FVFC-STRL fluctuates and increases slowly. It is verified that the proposed MTRL training framework effectively leverages the similarities among subtasks to enhance sample efficiency by more than 50%.

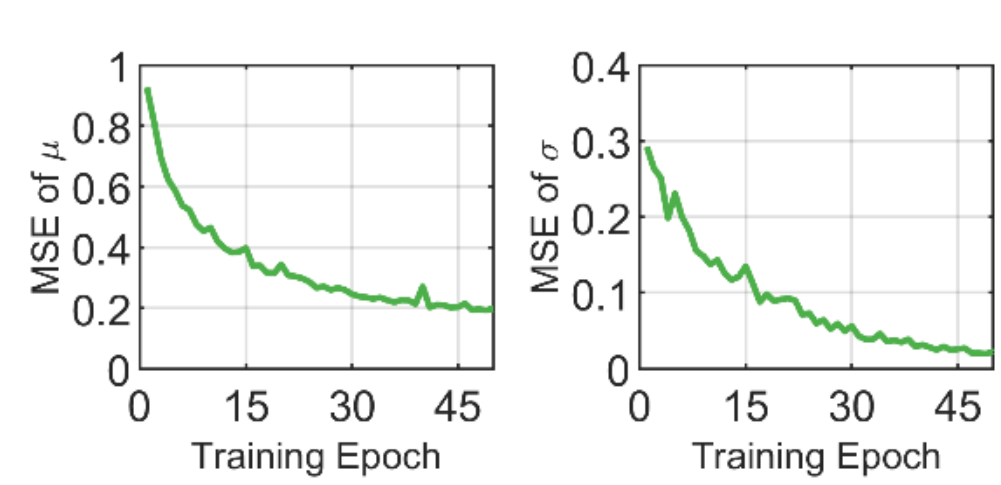

**Fig. 8** Policy distillation training process

In policy distillation process, the action space is normalized to [-1, 1], and an early stopping strategy is employed to prevent overfitting. The Mean Square Error (MSE) of the predicted mean ($\mu$) and standard deviation ($\sigma$) on the testing dataset during distillation process is presented in Fig. 8. The steady decrease of MSE indicates that the student network effectively learns decision-making behaviors consistent with those of the multiple teacher strategies.

**Table 3 Assembly performance in four subtasks in experiment.**

| Algorithm | Robustness | Performance in different subtasks (reward / success rate) | | | | Average performance |
|---|---|---|---|---|---|---|
| | | E-(1) | E-(2) | E-(3) | E-(4) | |
| FBCC | √ constant parameters | -2.2/ 30% | 5.0/ 54% | 4.9/ 54% | -1.1/ 30% | 1.7/ 42.0% |
| MDRL | × four task-specific strategies | 1.3/ 40% | 8.6/ 68% | 11.3/ 68% | 1.8/ 40% | 5.8/ 54.0% |
| FVFC-CP | √ constant parameters | 6.9/ 62% | 9.6/ 62% | 11.7/ 68% | 6.2/ 52% | 8.6/ 61.0% |
| FVFC-MTRL | × four task-specific strategies | 19.1/ 92% | 22.0/ 98% | 22.2/ 98% | 23.0/ 100% | 21.6/ 97.0% |
| **FVFC-MTRL-PD** | **√ one robust strategy** | **21.3/ 98%** | **22.6/ 100%** | **22.2/ 98%** | **22.3/ 98%** | **22.1/ 98.5%** |

## 4.3. Compliance and robustness verification

The constructed strategy, FVFC-MTRL-PD, is evaluated over 50 trials in each of the four subtasks and compared against other baseline methods. The average rewards and success rates are summarized in Table 3. FBCC and FVFC-CP adopt constant controller parameters for all subtasks. FVFC-MTRL-PD is the proposed strategy that is robust to all subtasks. These three strategies are considered robust, as they do not require prior knowledge of which specific subtask is currently being executed. In contrast, MDRL and FVFC-MTRL are non-robust strategies; thus, each subtask is tested using the strategy specifically trained for that subtask. Non-robust policies typically lack cross-task generalization capability, exhibiting significant performance degradation when directly transferred to different task configurations. For example, when applying the FVFC-MTRL policy trained in E-(4) to subtask E-(1), the average success rate decreases to 82% and the average control steps increase by 25%. As shown in Table 3, FVFC-MTRL demonstrates strong performance in each specific subtask, and FVFC-MTRL-PD further maintains the high performance of FVFC-MTRL while additionally achieving robustness across different subtasks. It outperforms all other methods, achieving the highest average reward of 22.1 and success rate of 98.5%.

To quantitatively evaluate the assembly compliance, the maximum forces/moments, and the number of control steps are recorded for each successful assembly over 50 trials per subtask. The average values for subtasks $T_2$ and $T_4$ are reported in Tables 4 and 5, respectively. It is verified that FVFC-MTRL-PD achieves the best assembly force compliance and the highest assembly efficiency among all evaluated strategies.

**Table 4 Maximum contact force and control step in subtask $T_2$.**

| Algorithm | Maximum force/ N | | | Maximum moment/ Nm | | | Control step |
|---|---|---|---|---|---|---|---|
| | $F_x$ | $F_y$ | $F_z$ | $M_x$ | $M_y$ | $M_z$ | |
| FBCC | 0.656 | 0.469 | 2.129 | 0.024 | 0.036 | **0.002** | 96.9 |
| MDRL | 0.581 | 0.402 | 1.982 | 0.021 | 0.032 | **0.002** | 82.6 |
| FVFC-CP | 0.349 | 0.231 | 1.214 | **0.011** | 0.020 | **0.002** | 88.9 |
| Ours* | **0.207** | **0.193** | **1.028** | **0.011** | **0.012** | **0.002** | **81.0** |

Ours*: FVFC-MTRL-PD

**Table 5 Maximum contact force and control step in subtask $T_4$.**

| Algorithm | Maximum force/ N | | | Maximum moment/ Nm | | | Control step |
|---|---|---|---|---|---|---|---|
| | $F_x$ | $F_y$ | $F_z$ | $M_x$ | $M_y$ | $M_z$ | |
| FBCC | 0.715 | 0.487 | 1.818 | 0.025 | 0.040 | **0.002** | 85.5 |
| MDRL | 0.590 | 0.358 | 1.394 | 0.017 | 0.033 | **0.002** | 86.5 |
| FVFC-CP | 0.315 | 0.271 | 0.468 | 0.013 | 0.017 | **0.002** | 81.2 |
| Ours* | **0.182** | **0.201** | **0.202** | **0.012** | **0.010** | **0.002** | **80.1** |

Ours*: FVFC-MTRL-PD

To more clearly illustrate the contact forces during the assembly process, the force trajectories from a single assembly trial under identical initial pose errors in subtask $T_2$ and $T_3$ are plotted in Fig. 9. When using FVFC-MTRL-PD, no significant jamming is observed during the insertion, and the contact forces are substantially reduced. For subtask $T_2$, as the insertion depth increases, the axial force $F_z$ increases steadily due to friction caused by the interference fit. Compared with the force control-based strategies (FBCC and MDRL), the FVFC method improves the assembly force compliance. Furthermore, the trained RL strategy further optimizes the control performance by better configuration of the FVFC controller parameters.

## 4.4. Batch precision assembly verification

To sufficiently verify the robustness of FVFC-MTRL-PD for the batch precision assembly task with uncertain fit amounts (-0.04 mm~0.04 mm), eight groups of components (G1~G8) are randomly selected from the batch-manufactured parts shown in Fig. 6 (c) for batch testing. Groups G1 and G2 exhibit clearance fits; thus, MDRL trained in subtask $T_4$ is selected for comparison. Groups G3~G8 exhibit interference fits, so MDRL trained in subtask $T_1$ is used for comparison. Notably, if the MDRL strategy trained on clearance-fit components is applied to interference-fit assembly tasks, it fails to complete the task in most cases, demonstrating its lack of generalization capability.

Each group of components is tested over 50 assembly trials with random initial alignment pose errors between pegs and holes, and the average rewards and success rates are shown in Table 6. The interference force reported in Table 6 implicitly corresponds to the axial force $F_z$ when the insertion is successfully completed, serving as an indirect indicator of the fit amount. During the MTRL training process in Section 4.2.2, the interference forces in $T_1$ and $T_2$ are approximately 3N and 1N, respectively.

In the batch assembly tasks, FVFC-MTRL-PD consistently maintains superior performance. In contrast, MDRL exhibits lower and less stable control performance even though the strategy trained in the most similar subtask is employed. For Group G6, the interference force exceeds the range of four subtasks during training, demonstrating that FVFC-MTRL-PD is capable of generalizing to unseen fit amounts at a certain extent. Furthermore, Groups G7 and G8 represent two intentionally designed, more challenging assembly scenarios, featuring hole with non-chamfered edges and pegs with manufacturing-induced deformations. For these groups, the MDRL approach failed completely (0% success rate), whereas the proposed method still achieved success rates of approximately 90%.

To illustrate the force compliance in the batch precision assembly task, five assembly trials are selected from each of the eight experimental groups, and the corresponding force heatmaps are presented in Fig. 10. The peak values of the contact forces $F_x$ and $F_y$ remain below 0.7 N during the batch precision assembly tasks. For clearance-fit assembly, the axial force $F_z$ remains low throughout the assembly process, whereas in interference-fit

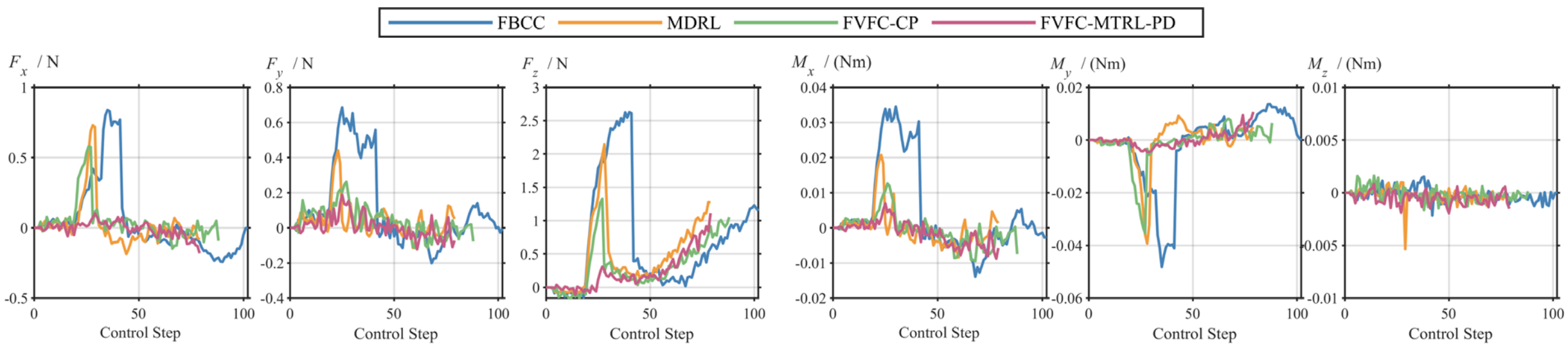

**(a) Force trajectories during once assembly process in subtask $T_2$**

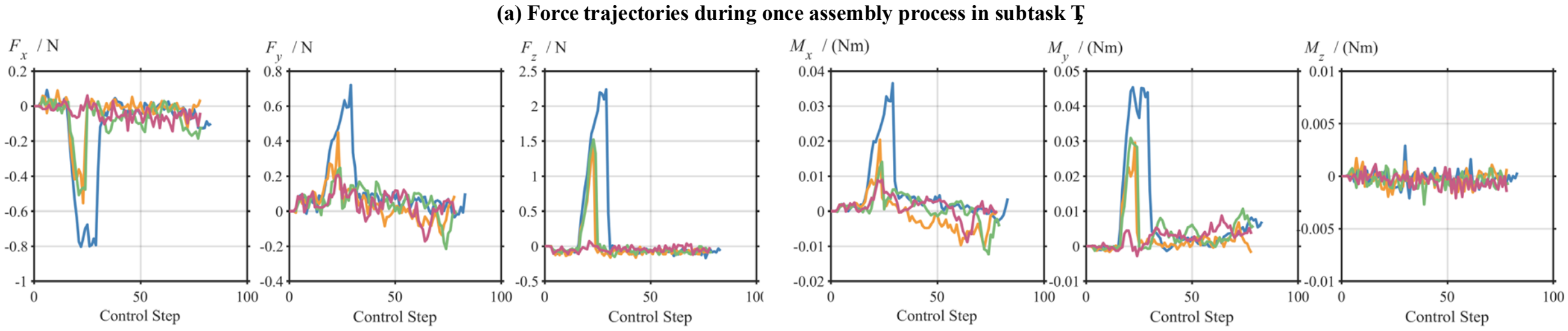

**(b) Force trajectories during once assembly process in subtask $T_3$**

**Fig. 9.** Comparisons of the force trajectories of different control strategies under the same initial pose errors.

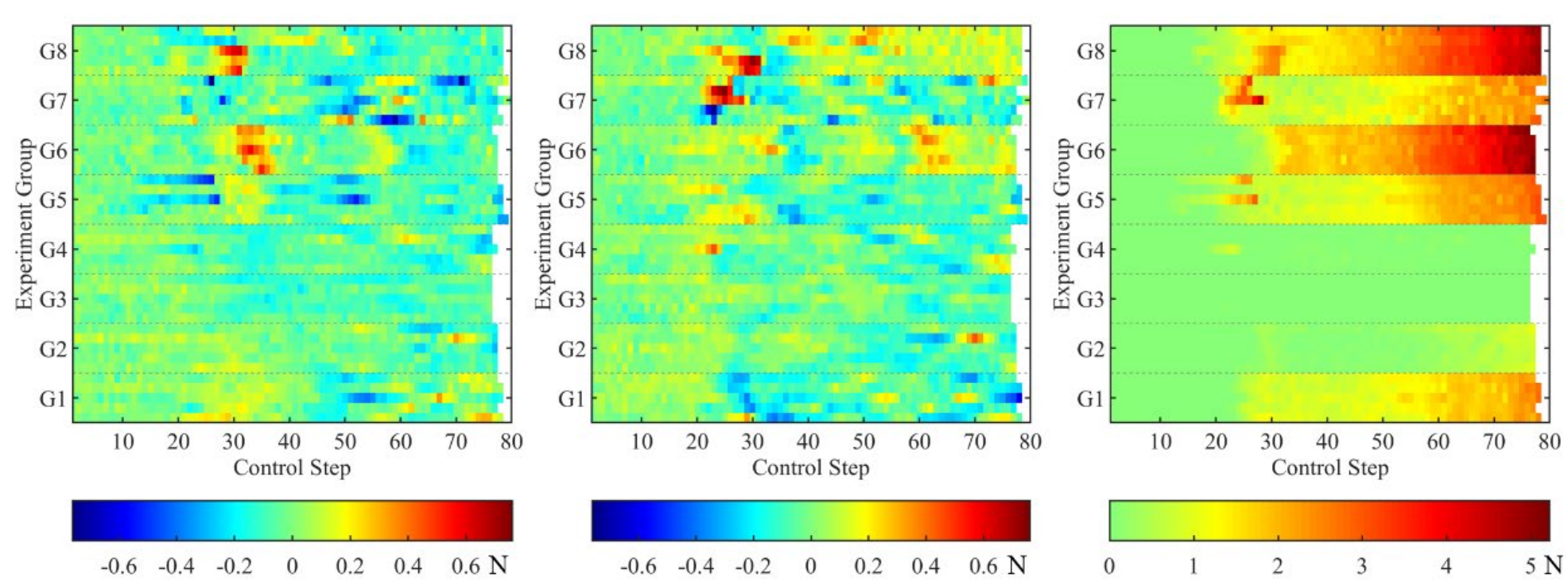

**Fig. 10.** Force trajectories in 40 trials (5 for each group) in the batch precision assembly task by using FVFC-MTRL-PD.

cases, $F_z$ increases linearly and continuously after the peg successfully enters the hole. FVFC-MTRL-PD exhibits stable and low contact forces across all eight component groups, verifying its superior force compliance and robustness.

**Table 6 Performance in batch precision assembly experiments.**

| Group id | Fit type (interference force) | Performance (reward/success rate) | |
|---|---|---|---|
| | | FVFC-MTRL-PD | MDRL ($T_1$ or $T_4$) |
| G1 | Clearance ( / ) | **22.8/ 100%** | 4.9/ 76% |
| G2 | Clearance ( / ) | **22.7/ 100%** | 11.8/ 82% |
| G3 | Interference (2.67 N) | **20.0/ 98%** | -4.2/ 22% |
| G4 | Interference (0.92 N) | **22.1/ 100%** | 0.3/ 36% |
| G5 | Interference (3.03 N) | **20.3/ 96%** | 6.4/ 52% |
| G6 | Interference (5.18 N) | **21.6/ 100%** | 10.3/ 68% |
| G7* | Interference (2.57 N) | **18.0/ 92%** | -9.5/ 0% |
| G8* | Interference (4.90 N) | **16.9/ 88%** | -9.9/ 0% |

## 5. Conclusion

The main contributions of this paper are as follows. For robotic batch assembly, an efficient framework for constructing the robust and compliant assembly control strategy is proposed, which consists of task decomposition, subtask reinforcement learning, and strategy integration. The FVFC-driven RL adopts a well-designed force-vision fusion controller and expanded state representations to improve the control performance of the final control strategy. The MTRL training framework improves 50% training efficiency for multiple subtasks by capturing the implicit shared feature representations. Finally, multi-teacher policy distillation integrates the capabilities and experiences acquired from multiple subtasks, contributing to the development of the final robust control strategy.

Unlike most existing studies that focus on learning the assembly strategy for a specific group of peg and hole, this paper focuses on the control robustness and compliance in the robotic batch precision assembly task, which more closely corresponds to the industrial high-precision production scenarios. The primary objective of this paper is to develop a control strategy that is robust to variations in fit types and fit amounts. From a broader perspective, the proposed framework is generalizable and can also be used to construct the assembly strategy that is robust to geometric variations, such as different component shapes (Details can be found in the supplementary materials).

In the future, how to accurately simulate precision assembly processes (especially interference-fit assembly), and effectively transfer the learned knowledge from simulation to real-world applications are more attractive and important research directions.

**Acknowledgements**

This study was financially supported by the National Nature Science Foundation of China [Grant No. 52375019].